\documentclass[runningheads]{llncs}
\usepackage[T1]{fontenc}
\usepackage{graphicx}
\usepackage{amsmath}
\usepackage{amssymb, amsfonts}
\usepackage{amsmath,amssymb}
\usepackage{makecell}
\usepackage{booktabs}
\usepackage{rotating}
\usepackage{array}
\usepackage{multirow}
\usepackage{comment}
\usepackage{diagbox}
\usepackage{epsfig}
\usepackage[linesnumbered,ruled,vlined]{algorithm2e}
\usepackage[dvipsnames]{xcolor,colortbl}
\usepackage[final]{microtype}
\usepackage{lipsum}
\usepackage{appendix}

\DeclareMathOperator{\E}{\mathbb{E}}

\newcommand{\AMAE}{\textsc{AMAE}}

\usepackage{tikz}
\usepackage{scalerel}

\usepackage{hyperref}
\hypersetup{
    colorlinks   = true,
    citecolor    = green
}

\begin{document}
\title{AMAE: Adaptation of Pre-Trained Masked Autoencoder for Dual-Distribution Anomaly Detection in Chest X-Rays}
\titlerunning{AMAE}
%


\author{Behzad Bozorgtabar\inst{1,3}\thanks{CONTACT Behzad Bozorgtabar. Email: behzad.bozorgtabar@epfl.ch.} 
\and 
Dwarikanath Mahapatra\inst{2}
\and 
Jean-Philippe Thiran\inst{1,3}}
\authorrunning{B. Bozorgtabar et al.}

\institute{École Polytechnique Fédérale de Lausanne (EPFL), Lausanne, Switzerland
\and
Inception Institute of AI (IIAI), Abu Dhabi, UAE
\and
Lausanne University Hospital (CHUV), Lausanne, Switzerland\\
}

%
%
%

\newcommand{\Thomas}[1]{\textcolor{blue}{#1}}
\newcommand{\Behzad}[1]{\textcolor{red}{#1}}

\newcommand{\aug}{\tilde{\boldsymbol{x}}}
\newcommand{\im}{\boldsymbol{x}}
\newcommand{\glob}{\bar{\boldsymbol{z}}}
\newcommand{\cls}{\texttt{[CLS]}}
\newcommand{\model}{g}
\newcommand{\backbone}{f}
\newcommand{\head}{h}
\newcommand{\bag}{\boldsymbol{X}}
\newcommand{\instance}{\boldsymbol{x}}
\newcommand{\ilabel}{y}
\newcommand{\blabel}{Y}

\newcommand{\vit}{\text{ViT}}
\newcommand{\resnet}{\text{ResNet}}
\newcommand{\abmil}{\text{AbMIL}}
\newcommand{\transmil}{\text{TransMIL}}
\newcommand{\clam}{\text{CLAM}}
\newcommand{\dino}{\text{DINO}}
\newcommand{\overbar}[1]{\mkern 1.5mu\overline{\mkern-1.5mu#1\mkern-1.5mu}\mkern 1.5mu}
\newcommand{\supp}{\textbf{Supplementary Material}}

\newcommand{\ck}{\textcolor{green!80!black}{\ding{51}}}
\newcommand{\xk}{\textcolor{red}{\ding{55}}}
\newcommand*\rot{\rotatebox{90}}

\definecolor{light_gray}{rgb}{0.85,0.85,0.85}
\newcolumntype{g}{>{\columncolor{light_gray}}c}
\maketitle              
%

\begin{abstract}
Unsupervised anomaly detection in medical images such as chest radiographs is stepping into the spotlight as it mitigates the scarcity of the labor-intensive and costly expert annotation of anomaly data. However, nearly all existing methods are formulated as a one-class classification trained only on representations from the normal class and discard a potentially significant portion of the unlabeled data. This paper focuses on a more practical setting, dual distribution anomaly detection for chest X-rays, using the entire training data, including both normal and unlabeled images. Inspired by a modern self-supervised vision transformer model trained using partial image inputs to reconstruct missing image regions– we propose $\AMAE$, a two-stage algorithm for adaptation of the pre-trained masked autoencoder (MAE). Starting from MAE initialization, $\AMAE$ first creates synthetic anomalies from only normal training images and trains a lightweight classifier on frozen transformer features. Subsequently, we propose an adaptation strategy to leverage unlabeled images containing anomalies. The adaptation scheme is accomplished by assigning pseudo-labels to unlabeled images and using two separate MAE based modules to model the normative and anomalous distributions of pseudo-labeled images. The effectiveness of the proposed adaptation strategy is evaluated with different anomaly ratios in an unlabeled training set. $\AMAE$ leads to consistent performance gains over competing self-supervised and dual distribution anomaly detection methods, setting the new state-of-the-art on three public chest X-ray benchmarks - RSNA, NIH-CXR, and VinDr-CXR.

\end{abstract}
%
%
\section{Introduction}
To reduce radiologists’ reading burden and make the diagnostic process more manageable, especially when the number of experts is scanty, computer-aided diagnosis (CAD) systems, particularly deep learning-based anomaly detection~\cite{bozorgtabar2022anomaly,pang2021deep,spahr2021self,bozorgtabar2023attention,bozorgtabar2020salad,bozorgtabar2021sood}, have witnessed the flourish due to their capability to detect rare anomalies for different imaging modalities including chest X-ray (CXR). Nonetheless, unsupervised anomaly detection methods~\cite{tian2021constrained,schlegl2019f} are strongly preferred due to the difficulties of highly class-imbalanced learning and the tedious annotation of anomaly data for developing such systems. Most current anomaly detection methods are formulated as a one-class classification (OCC) problem~\cite{ruff2018deep}, where the goal is to model the distribution of normal images used for training and thus detect abnormal cases that deviate from normal class at test time. On this basis, image reconstruction based, e.g., autoencoder~\cite{gong2019memorizing} or GANs~\cite{schlegl2019f,goodfellow2014generative}, self-supervised learning (SSL) based, e.g., contrastive learning~\cite{tian2021constrained}, and embedding-similarity-based methods~\cite{defard2021padim} have been proposed for anomaly detection. Some recent self-supervised methods proposed synthetic anomalies using cut-and-paste data augmentation~\cite{li2021cutpaste,sato2022anatomy} to approximate real sub-image anomalies. Nonetheless, their performances lag due to the lack of real anomaly data. More importantly, these methods have often ignored readily available unlabeled images. More recently, similar to our method, DDAD~\cite{cai2022dual} leverages readily available unlabeled images for anomaly detection, but it requires training an ensemble of several reconstruction-based networks. Self-supervised model adaptation on \textit{unlabeled data} has been widely investigated using convolutional neural networks (CNNs) in many vision tasks via self-training~\cite{prabhu2021sentry}, contrastive learning~\cite{tian2021constrained,spahr2021self}, and anatomical visual words~\cite{haghighi2021transferable}. Nonetheless, the adaptation of vision transformer (ViT)~\cite{dosovitskiy2020image} architectures largely remains unexplored, particularly for anomaly detection. Recently, masked autoencoder (MAE)~\cite{he2022masked} based models demonstrated great scalability and substantially improved several self-supervised learning benchmarks~\cite{xiao2023delving}. 
\\
In this paper, inspired by the success of the MAE approach, we propose a two-stage algorithm for ``\textbf{A}daptation of pre-trained \textbf{M}asked \textbf{A}uto\textbf{E}ncoder'' (\textbf{$\AMAE$}) to leverage simultaneously normal and unlabeled images for anomaly detection in chest X-rays. As for \textbf{Stage 1} of our method, (i) $\AMAE$ creates synthetic anomalies from only normal training images, and the usefulness of pre-trained MAE~\cite{he2022masked} is evaluated by training a lightweight classifier using a proxy task to detect synthetic anomalies.  (ii) For the \textbf{Stage 2}, $\AMAE$ customizes the recipe of MAE adaptation based on an unlabeled training set. In particular, we propose an adaptation strategy based on reconstructing the masked-out input images. The rationale behind the proposed adaptation strategy is to assign pseudo-labels to unlabeled images and train two separate modules to measure the distribution discrepancy between normal and pseudo-labeled abnormal images. (iii) We conduct extensive experiments across three chest X-ray datasets and verify the effectiveness of our adaptation strategy in apprehending anomalous features from unlabeled images. In addition, we evaluate the model with different anomaly ratios (ARs) in an unlabeled training set and show consistent performance improvement with increasing AR.



\begin{figure*}[t!]
    \includegraphics[width=1\textwidth]
    {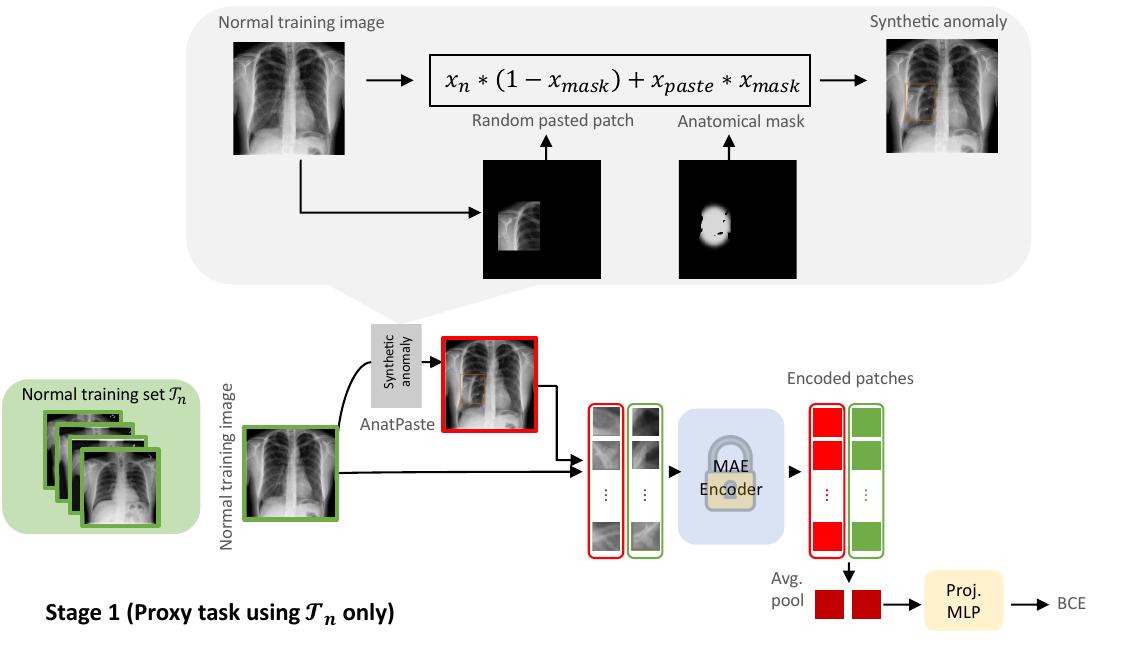}
    \caption{\textbf{Schematic overview of AMAE training (Stage 1)}. \textbf{Top}. Illustration of Anatpaste augmentation~\cite{sato2022anatomy} generated from normal training images.  \textbf{Bottom}. Starting from MAE initialization, only the MLP-based projection head (\texttt{Proj.}) is trained to classify synthetic anomalies.} 
    \vspace{-1.5em}
    \label{fig1}
\end{figure*}

\section{Method}
\noindent
\textbf{Notation.} We first formally define the problem setting for the proposed dual-distribution anomaly detection. Contrary to previous unsupervised anomaly detection methods, $\AMAE$ fully uses unlabeled images, yielding a training data $\mathcal{T}_{train}=\mathcal{T}_{n}\cup \mathcal{T}_{u}$ consisting of both normal $\mathcal{T}_{n}$ and unlabeled $\mathcal{T}_{u}$ training sets. We denote the normal training set as $\mathcal{T}_{n}=\left \{ \instance_{ni} \right \}_{i=1}^{N}$, with $N$ normal images, and the unlabeled training set as $\mathcal{T}_{u}=\left \{ \instance_{ui} \right \}_{i=1}^{M}$, with $M$ unlabeled images to be composed of both normal and abnormal images. At test time, given a test set $\mathcal{T}_{test}=\left \{ \left ( \instance_{ti},\ilabel_{i} \right ) \right \}_{i=1}^{S}$ with $S$ normal or abnormal images, where $\ilabel_{i}\in \left \{ 0,1 \right \}$ is the corresponding label to $\instance_{ti}$ (0 for normal (negative) and 1 for abnormal (positive) image), the trained anomaly detection model should identify whether the test image is abnormal or not. 

\vspace{1em}
\noindent
\textbf{Architecture.} Our architecture is $\blacktriangleright\mathrel{\mkern-4mu}<$-shaped: the ViT-small (ViT-S/16) \cite{dosovitskiy2020image} encoder $f$ followed by a self-supervised ViT head $g$ and lightweight (3-layer) multilayer perception (MLP) anomaly classifier (projection head) $h$, simultaneously. Starting from the pre-trained MAE on 0.3M unlabeled chest X-rays and officially released \href{https://github.com/lambert-x/medical_mae}{checkpoints}, we use exactly the same ViT encoder $f$ and decoder $g$ as MAE~\cite{xiao2023delving}.





\subsection{Stage 1- Proxy Task to Detect Synthetic Anomalies} 
$\AMAE$ starts the first training stage using only normal training images by defining a proxy task to detect synthetic anomalies shorn of real known abnormal images. For this purpose, we utilize the state-of-the-art (SOTA) anatomy-aware cut-and-paste augmentation, AnatPaste~\cite{sato2022anatomy}, to create synthetic anomalies from only a set of normal training images $\mathcal{T}_{n}$. AnatPaste integrates an anatomical mask $\instance_{mask}$ created from unsupervised lung region segmentation, which guides generating anomalous images via cutting a patch from a normal chest radiograph $\instance_{n}$ and randomly pasting it at another image location $\instance_{paste}$ as:
\begin{align} 
\label{eq:1}
    \mathbf{Aug}\left (\instance_{n}  \right )=\instance_{n}\ast \left ( 1-\instance_{mask} \right )+\instance_{paste}\ast \instance_{mask} 
\end{align}
where $\mathbf{Aug\left ( \cdot  \right )}$ is the AnatPaste augmentation (see \cite{sato2022anatomy} for more details).



Given a normal training set $\mathcal{T}_{n}$, for each normal image $\instance_{n}\sim \mathcal{T}_{n}$, we create a synthetic anomaly, denoted as $\mathbf{Aug}\left (\instance_{i}  \right )$. In preparation for input to the frozen ViT encoder, $f_{0}$ (obtained by MAE pre-training), each input image with the  $h\times w$ spatial resolution is split into $T=\left ( h/p \right )\times \left ( w/p \right )$ patches of size $p \times p$. Then, for every input patch, a token is created by linear projection with an added positional embedding. The sequence of tokens is then fed to the frozen ViT encoder $f_{0}$ consisting stack of transformer blocks, yielding the embeddings of tokens $\boldsymbol{z}_{i}^{1},\boldsymbol{z}_{i}^{2}\in \mathbb{R}^{T\times d}$ corresponding to $i^{th}$ normal and synthetic anomaly images. The returned embeddings $\boldsymbol{z}_{i}^{1},\boldsymbol{z}_{i}^{2}\in \mathbb{R}^{T\times d}$ are pooled via \textit{average pooling} to form $d$-dimensional embeddings, which are fed to an MLP anomaly classifier projection head $h$ (see Fig. \ref{fig1} for schematic overview). Subsequently, we only train an anomaly classifier projection head $h$ on top of the frozen embeddings to detect synthetic anomalies using the cross-entropy loss $l_{ce}$ as follows:
\begin{align} 
\label{eq:1}
    h_{0}=\arg \min_{h}\E_{\instance_{n}\sim  \mathcal{T}_{n}}\left [l_{ce}\left ( h \circ f_{0}\left ( \instance_{n} \right ),0 \right )+ l_{ce}\left ( h \circ f_{0}\left ( \mathbf{Aug}\left (\instance_{n}  \right )
 \right ),1 \right )  \right ]
\end{align}
We set the label for the normal image to 0 and 1 otherwise (synthetic anomaly). The above gradient-based optimization produces a trained classifier projection head $h_{0}$. Thus, the whole architecture can be trained with much fewer parameters while making only the classifier projection head specialized at recognizing anomalies without influencing the ViT encoder.

\begin{figure*}[t!]
    \includegraphics[width=1\textwidth]{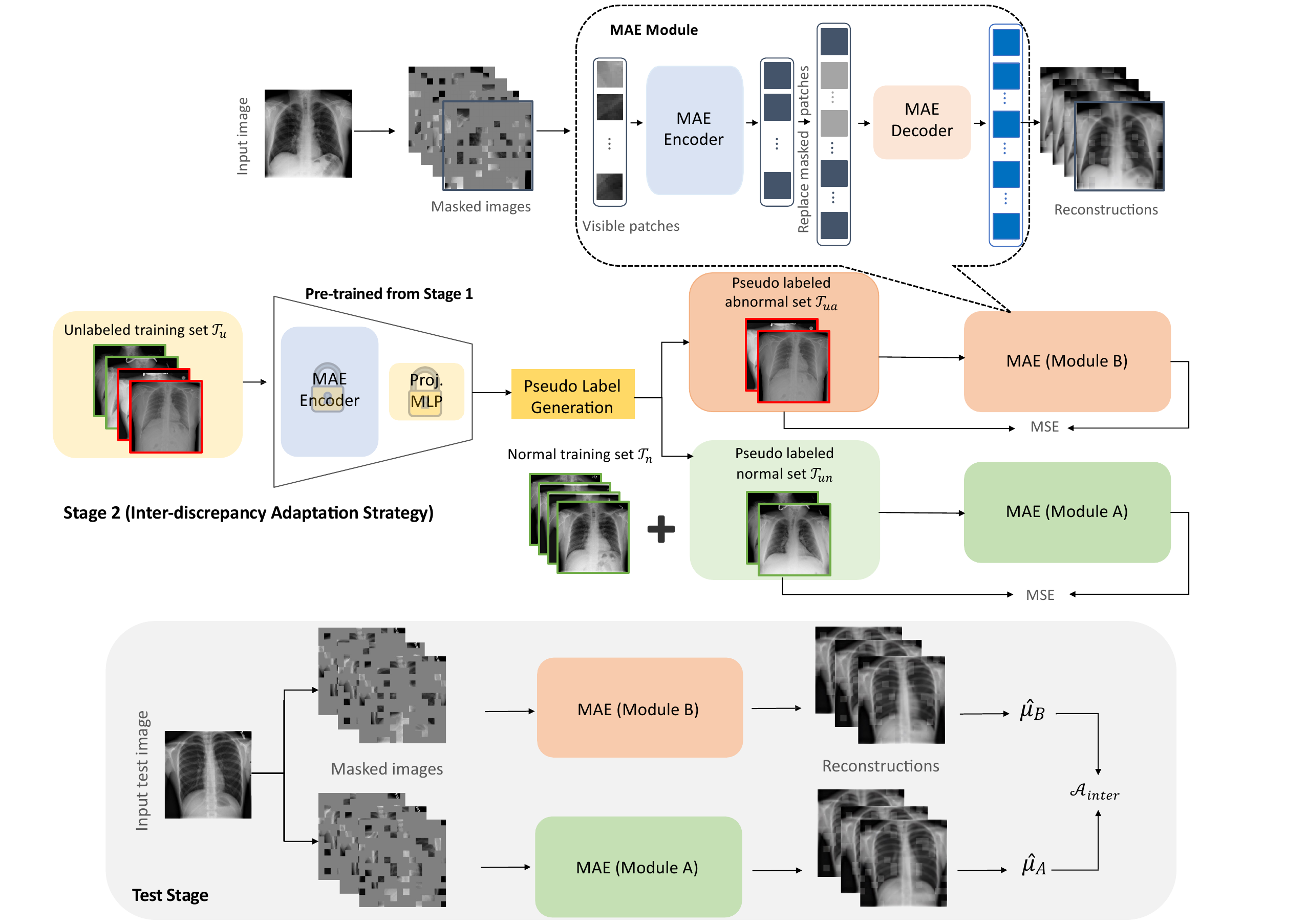}
    \caption{\textbf{Schematic overview of $\AMAE$ training (Stage 2) and test stage}. \textbf{Top}. Starting from MAE initialization, our adaptation strategy first assigns pseudo labels to unlabeled images using a pre-trained classifier from \textbf{Stage 1} and uses two separate MAE modules to model the normative and anomalous distributions of pseudo-labeled images. \textbf{Bottom}. At test time, the discrepancy between the average of multiple reconstructions from two modules is used to compute the anomaly score.}
    \vspace{-1.5em}
    \label{fig2}
\end{figure*}

\subsection{Stage 2- MAE Inter-Discrepancy Adaptation}
The proposed MAE adaptation scheme is inspired by \cite{cai2022dual} to model the dual distribution of training data. Unlike \cite{cai2022dual}, which treats all unlabeled images similarly, we propose assigning pseudo-labels to unlabeled images and formulating anomaly detection by measuring the distribution discrepancy between normal and pseudo-labeled abnormal images from unlabeled sets. We use a pre-trained anomaly classifier (\textbf{Stage 1}) to assign pseudo labels to unlabeled images. To begin with, for each unlabeled image $\instance_{u}\sim \mathcal{T}_{u}$, we consider the anomaly detection model’s confidence from \textbf{Stage 1} ($h_{0} \circ f_{0}\left ( \instance_{u} \right )$). Those images on which the model is highly confident (normal or abnormal) are treated as reliable images used for adaptation. For this purpose, we collect all output probabilities and opt for a threshold per class $t_{c}$ corresponding to each class's top $K$-th percentile ($K$=50) of all given confidence values. Those unlabeled images deemed reliable yield two subsets: a subset of pseudo-labeled normal images $\mathcal{T}_{un}$ and a subset of pseudo-labeled abnormal images $\mathcal{T}_{ua}$. We then utilize two MAE-based modules, \textbf{Module A} and \textbf{Module B} (see Fig. \ref{fig2}, \textbf{Top}), using the same MAE architecture and pixel-wise mean squared error (MSE) optimization in \cite{he2022masked}. Within each module, the input patches for each image are masked out using a set of $L$ random masks $\left \{\boldsymbol{m}^{\left ( j \right )}\in \left \{ 0,1 \right \} ^{T} \right \}_{j=1}^{L}
$ in which a different small subset of the patches (ratio of 25\%) is retained each time to be fed to the ViT encoder $f$. The lightweight ViT decoder $g$ receives unmasked patches' embeddings and adds learnable masked tokens to replace the masked-out patches. Subsequently, the full set of embeddings of visible patches and masked tokens with added positional embeddings to all tokens is processed by the ViT decoder $g$ to reconstruct the missing patches of each image in pixels. This yields the reconstructed image $\hat{\instance}^{\left ( j \right )}=g \circ f\left (\boldsymbol{m}^{\left ( j \right )}\left ( \instance \right )  \right )$, which is then compared against the input image $\instance$ to optimize both ViT encoder $f$ and decoder $g$:

\begin{align} 
\label{eq:2}
    f^{\ast },g^{\ast }=\arg \min_{f,g}\mathbb{E}_{\instance\sim  \mathcal{T}_{train}}\left [\frac{1}{L}\sum_{j=1}^{L}l_{mse}\left (\boldsymbol{m}^{\left ( j \right )}\left (\hat{\instance}^{\left ( j \right )}  \right ), \boldsymbol{m}^{\left ( j \right )}\left (\instance  \right )  \right )  \right ]
\end{align}
All pixels in the $t^{th}$ patch of both input image and reconstructed images are multiplied by $\left (\boldsymbol{m}^{\left ( j \right )}  \right )_{t}\in \left \{ 0,1 \right \}$. The above self-supervised loss term averages $L$ pixel-wise mean squared errors for each image. \textbf{Module A} is trained on a combination of the normal training set $\mathcal{T}_{n}$ and pseudo-labeled normal images from the unlabeled set $\mathcal{T}_{un}$. In contrast, \textbf{Module B} is trained using only pseudo-labeled abnormal images from an unlabeled set $\mathcal{T}_{ua}$. Optimization for Eq. \ref{eq:2} always starts from pre-trained $f_{0}$ and $g_{0}$, and we reset the MAE weights to $f_{0}$ and $g_{0}$ before training each module. A high discrepancy between the reconstruction outputs of the two modules can indicate potential abnormal regions. 
Similar to the training stage, we apply $L$ random masks to the test image $\instance_{t}\sim \mathcal{T}_{test}$ to obtain $L$ reconstructions (see Fig. \ref{fig2}, \textbf{Bottom}). Thus, the anomaly score based on the inter-discrepancy of the two MAE modules is computed as follows:

\begin{figure*}[t!]
  \includegraphics[width=1\textwidth]{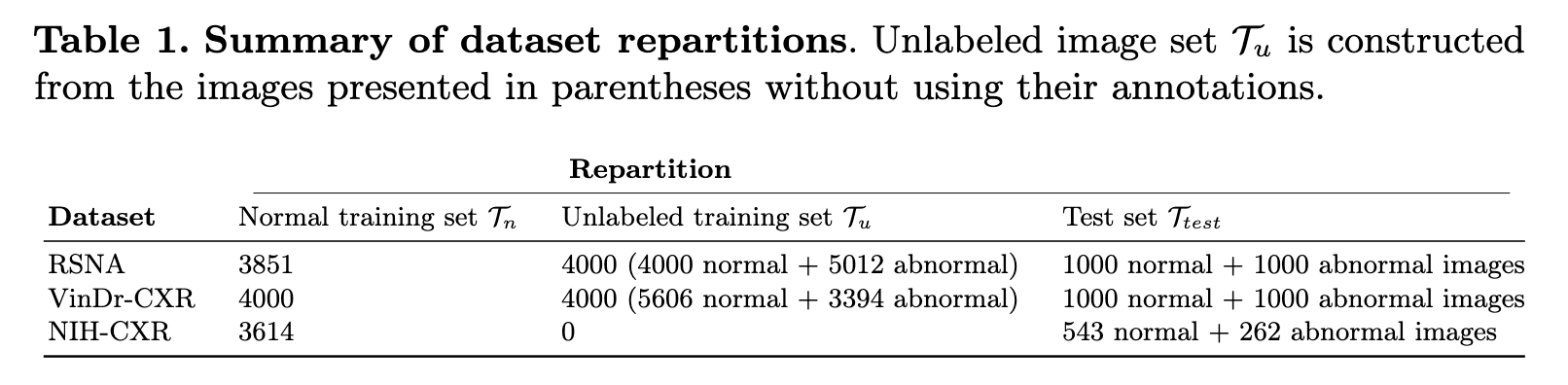}
    \vspace{-1.5em}
    \label{fig22}
\end{figure*}


\begin{align} 
\label{eq:3}
    \mathcal{A}_{inter}^{p}=\left | \hat{\mu}_{A}^{p}-\hat{\mu}_{B}^{p} \right |
\end{align}
where $p$ is the index of pixels, $\hat{\mu}_{A}$ and $\hat{\mu}_{B}$ are the \textit{mean} maps of $L$ reconstructed images from \textbf{Module A} and \textbf{Module B}, respectively. The pixel-level anomaly scores for each image are averaged, yielding the image-level anomaly score.

\section{Experiments}
\label{sec:Experiments}



\setcounter{table}{1} \renewcommand\thetable{\arabic{table}}
\begin{table}[t]
    \centering
    \caption{\textbf{Performance comparison with SOTA methods (Avg. over four replicate).} The two best results for each protocol are highlighted in \textbf{bold} and \underline{underlined}. Note that “IN” refers to “ImageNet-Pretrained,” and “e2e” refers to end-to-end training. The experimental results of competing methods$^{\dagger}$ are taken from \cite{cai2023dual}.}  
\resizebox{\textwidth}{!}{
    \begin{tabular}{@{}c|c|c|cccccc|cc|cc@{}}
    
    \toprule
    
       \multirow{3}[3]{*}{Method} &  \multirow{3}[3]{*}{\rotatebox[origin=c]{90}{Protocol}} & \multirow{3}[3]{*}{\rotatebox[origin=c]{90}{Taxonomy}} &\multicolumn{6}{c}{\textbf{CXR Datasest \& Metrics}}  \\
       
      \cmidrule(lr){4-9} 
       
      \multicolumn{1}{c|}{} &
       \multicolumn{1}{c|}{} &
      \multicolumn{1}{c}{} & \multicolumn{2}{|c}{\textbf{RSNA}} & \multicolumn{2}{c}{\textbf{VinDr-CXR}} & 
      \multicolumn{2}{c}{\textbf{NIH-CXR}} \\
      
      \cmidrule(lr){4-5} \cmidrule(lr){6-7} \cmidrule(lr){8-9} 
      
       & & &\multicolumn{1}{c}{AUC \%} & AP \%& AUC \%& AP \%& AUC \%& \multicolumn{1}{c}{AP} \% \\ 

    \midrule

    AE$^{\dagger}$ & N & Rec.& 66.9 & 66.1 & 55.9& 60.3& 70.0 & \multicolumn{1}{c}{65.4}  \\
    AE-U$^{\dagger}$~\cite{mao2020abnormality} & N & Rec.& \underline{86.7} & \underline{84.7} & 73.8& 72.8&91.0 & \multicolumn{1}{c}{83.2}  \\
    f-AnoGAN$^{\dagger}$~\cite{schlegl2019f}& N & Rec.& 79.8 & 75.6 & \textbf{76.3}& \textbf{74.8}&84.1 & \multicolumn{1}{c}{79.2}  \\
    IGD$^{\dagger}$~\cite{chen2022deep}& N & Rec.& 81.2 & 78.0 & 59.2& 58.7& 85.2 & \multicolumn{1}{c}{80.3}  \\
    
    DRAEM$^{\dagger}$~\cite{zavrtanik2021draem}& N&  Rec.+SSL& 62.3 & 61.6 &  63.0 & 68.3 & 65.5 & \multicolumn{1}{c}{62.4}  \\
    CutPaste$^{\text{e2e}}$$^{\dagger}$~\cite{schluter2022natural} & N&  SSL & 55.0 & 58.0 & 54.6 & 55.5 & 58.0 & \multicolumn{1}{c}{-}  \\
    AnatPaste~\cite{sato2022anatomy}& N&  SSL & 83.1 & 83.7 & 66.0 & 66.2 & 94.0 & \multicolumn{1}{c}{93.5}  \\
    FPI$^{\dagger}$~\cite{tan2020detecting} & N&  SSL & 47.6 & 55.7& 48.2 & 49.9& 70.5 & \multicolumn{1}{c}{-}\\
    PII$^{\dagger}$~\cite{tan2021detecting} & N&  SSL & 82.9 & 83.6& 65.9 & 65.8& 92.2 & \multicolumn{1}{c}{-}\\
    
    
    NSA$^{\dagger}$~\cite{schluter2022natural} & N&  SSL & 82.2 & 82.6 & 64.4 & 65.8 & 94.1 & \multicolumn{1}{c}{-}\\
    DDAD-AE~\cite{cai2022dual}  & N&  Rec. & 69.4 & \multicolumn{1}{c}{-} & 60.1 & \multicolumn{1}{c}{-} & 73.0 & \multicolumn{1}{c}{71.5}
    \\
    \midrule
    \textbf{$\AMAE$$^{\text{IN}}$ - Stage 1} & N & SSL & 83.3 & 83.8 & 65.9 & 66.0 & \underline{94.1} &  \multicolumn{1}{c}{\underline{93.7}} \\
    \textbf{$\AMAE$ - Stage 1} & N & SSL & \textbf{86.8} & \textbf{84.9} & \underline{74.2} & \underline{72.9} & \textbf{95.0} &  \multicolumn{1}{c}{\textbf{94.9}} \\
    
    \midrule
    CutPaste$^{\text{e2e}}$$^{\dagger}$~\cite{schluter2022natural} & Y&  SSL & 59.8 & 61.7 & 59.2 & 60.0 & \multicolumn{2}{c}{-}  \\
    AnatPaste~\cite{sato2022anatomy}& Y&  SSL & \underline{84.4} & \underline{85.5} & 67.1 & \underline{67.5} & \multicolumn{2}{c}{-}  \\
    FPI$^{\dagger}$~\cite{tan2020detecting} & Y&  SSL & 46.6 & 53.8& 47.4 & 49.4&  \multicolumn{2}{c}{-}\\
    PII$^{\dagger}$~\cite{tan2021detecting} & Y&  SSL & 84.3 & 85.4& 66.8 & 67.2&  \multicolumn{2}{c}{-}\\
    
    
    NSA$^{\dagger}$~\cite{schluter2022natural} & Y&  SSL & 84.2 & 84.3 & 64.4 & 64.8 &  \multicolumn{2}{c}{-}\\
    DDAD-AE~\cite{cai2022dual} ($\mathcal{A}_{iner}$)   & Y&  Rec. & 81.5 & 81.0 & \underline{71.0} & \multicolumn{1}{c}{-} & \multicolumn{2}{c}{-}
    \\
    \midrule
     
     \textbf{$\AMAE$ - Stage 2 ($\mathcal{A}_{inter}$ )} & Y&  Rec. & \textbf{91.4} & \textbf{91.7} & \textbf{86.1} & \textbf{84.5}
     &  \multicolumn{2}{c}{-} \\
    \bottomrule    
    \end{tabular}
    }
    \label{tab:results_classification}
    \vspace{-1em}
\end{table}

\noindent
\textbf{Datasets.}
We evaluated our method on three public CXR datasets: 1) the RSNA Pneumonia Detection Challenge dataset\footnote{https://www.kaggle.com/c/rsna-pneumonia-detection-challenge}, 2) the VinBigData Chest X-ray Abnormalities Detection Challenge dataset (VinDrCXR)\footnote{https://www.kaggle.com/c/vinbigdata-chest-xray-abnormalities-detection}~\cite{nguyen2022vindr}, and a subset of 3) the curated NIH dataset (NIH-CXR)\footnote{https://nihcc.app.box.com/v/ChestXray-NIHCC/file/371647823217}~\cite{tang2020automated,schluter2022natural}, by including only posteroanterior view images of both male and female patients aged over 18. We show a summary of each dataset's repartitions in Table \textcolor{red}{1}. Except for NIH, where we use only the normal set $\mathcal{T}_{n}$ (OCC setting), for the other two datasets, we utilize both $\mathcal{T}_{n}$ and $\mathcal{T}_{u}$ and exact repartition files from \cite{cai2022dual} for model training.


\noindent
\textbf{Implementation details.}
 We adopt AdamW~\cite{loshchilov2018decoupled} optimizer and set the learning rate ($lr$) and batch size to $2.5e\text{-}4$ and 16, where we linearly warm up the $lr$ for the first 20 epochs and decay it following a cosine schedule thereafter till 150 epochs. We follow the exact recipe as \cite{xiao2023delving} for other hyperparameters. The number of generated masks $L$ per image is set to 2 and 4 for the adaptation and test stages. We use PyTorch 1.9~\cite{paszke2019pytorch} and train each model on a single GeForce RTX 2080 Ti GPU. We use the area under the ROC curve (AUC) and average precision (AP) for the evaluation metrics.


\begin{figure*}[t!]
    \includegraphics[width=1\textwidth]{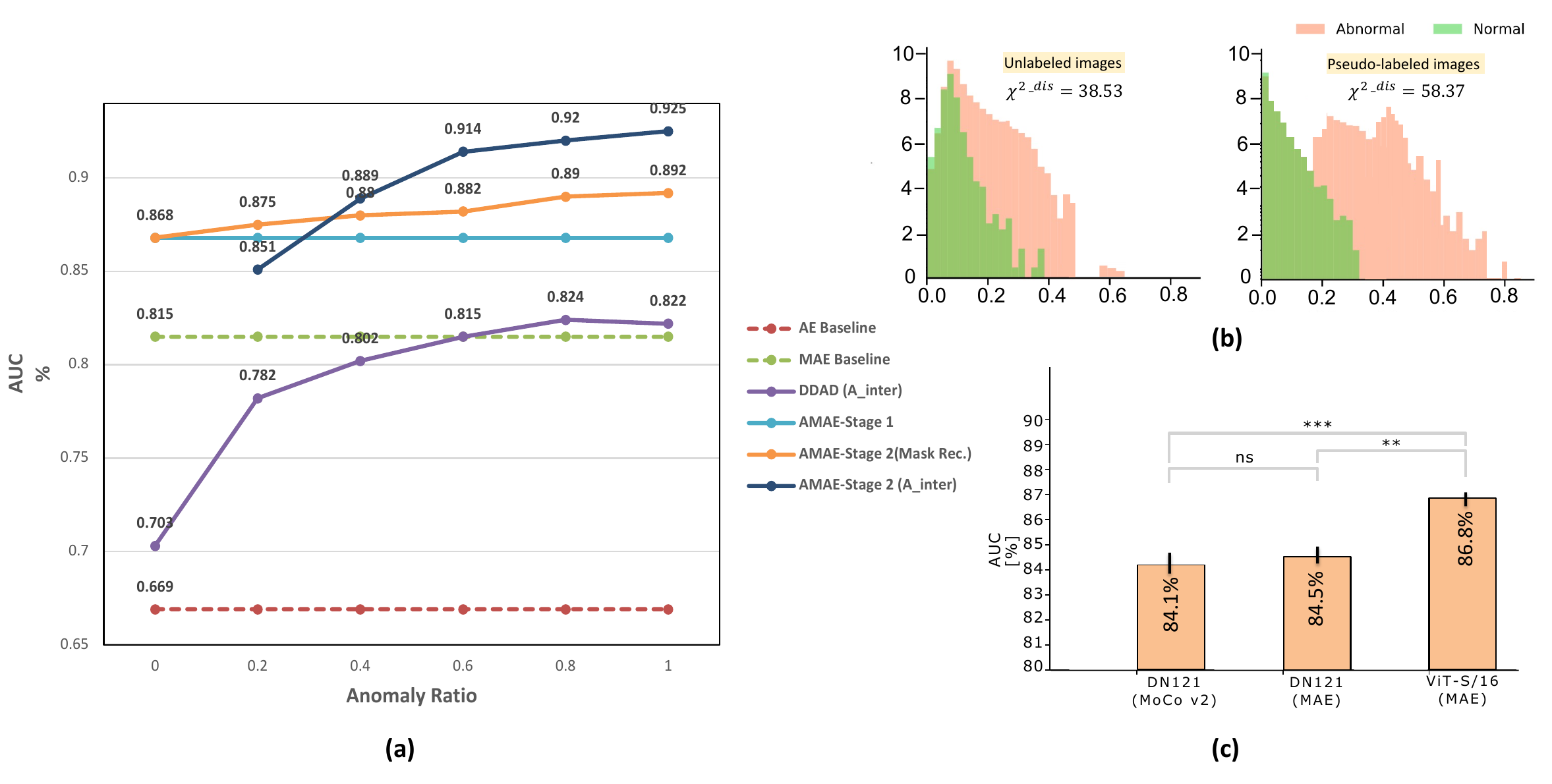}
    \caption{\textbf{Ablations on the RSNA dataset}. \textbf{(a)} Performance comparison with a varying AR of $\mathcal{T}_{u}$. \textbf{(b)} The $\chi ^{2-\text{distance}}$ of AS histograms \textit{with} and \textit{without} pseudo labeling. \textbf{(c)} Ablation for different backbones and pre-training schemes. Test AUC performances are presented as $mean\pm 1.96std$ averaged over four replicates. Samples are statistically tested for \emph{H\textsubscript{0}: means are similar}, using a bilateral Welch t-test. \emph{ns} non-significant, ** $pvalue \in [0.05, 0.01]$, *** $pvalue \in [0.01, 0.001]$.} 
    \vspace{-1.5em}
    \label{fig3}  
\end{figure*}

\vspace{0.5em}\noindent
{\textbf{Comparison with SOTA Methods.}} Table \textcolor{red}{2} compares $\AMAE$ with a comprehensive collection of SOTA methods, including self-supervised synthetic anomaly and reconstruction (Rec.) based methods using their official codes and under two experimental protocols. We use the \textbf{Y} protocol to indicate if access to the unlabeled images is possible in which an AR of 60\% of $\mathcal{T}_{u}$ is assumed in the experiments; otherwise, we use \textbf{N}. Under protocol \textbf{N} (OCC setting), except for VinDr-CXR, $\AMAE$-\textbf{Stage 1} achieves SOTA results on two CXR benchmarks, demonstrating the effectiveness of pre-trained ViT using MAE and synthetic anomalies. In particular, $\AMAE$-\textbf{Stage 1} surpasses the best-performing synthetic anomaly-based method, AnatPaste~\cite{sato2022anatomy}, with the same synthesis approach as ours but using ResNet18 as a feature extractor. Furthermore, outperforming MAE pre-training on ImageNet (e.g., improved AUC from 65.9\% to 74.2\% on the VinDr-CXR) indicates the importance of in-domain adaptation. Under the \textbf{Y} protocol, our adaptation strategy, $\AMAE$- \textbf{Stage 2} ($\mathcal{A}_{inter}$), outperforms the current SSL methods by a larger margin, underlining the importance of modeling the dual distribution and leveraging unlabeled images more effectively.



\vspace{0.5em}\noindent
{\textbf{Ablation Studies.}} To understand the effectiveness of $\AMAE$ in capturing abnormal features from unlabeled images, we conduct ablation experiments on the RSNA with the AR of $\mathcal{T}_{u}$ varying from 0 to 100\% (Fig. \ref{fig3} (\textbf{a})). Concerning reconstruction-based methods, the baseline aggregating multiple reconstructions via MAE achieves consistent improvement compared with the AE baseline (+14.6\% AUC), implying better capturing of fine-grained texture information. With an increasing AR of $\mathcal{T}_{u}$, our MAE adaptation strategy ($\AMAE$ - \textbf{Stage 2} ($\mathcal{A}_{inter}$ )) performs favorably against the SOTA method (DDAD~\cite{cai2022dual}), and $\AMAE$ without adaptation (\textbf{Stage 1}). Furthermore, We consider an additional baseline for model adaptation based on the patch reconstruction objective in \cite{he2022masked} on pooled normal and unlabeled images, denoted as ($\AMAE$ - \textbf{Stage 2} (Mask Rec. )). $\AMAE$ - \textbf{Stage 2} ($\mathcal{A}_{inter}$ ) rises with a steeper slope than $\AMAE$ - \textbf{Stage 2} (Mask Rec. ), e.g., improved AUC from 89\% to 92\% on AR=80\%, suggesting high-quality pseudo-labeled images. We also analyze the discriminative capability of our adaptation \textit{with} and \textit{without} pseudo labeling by levering all unlabeled images in \textbf{Module B}. We utilize the $\chi ^{2-\text{distance}}$ between the histograms of anomaly scores (AS) of normal and abnormal images in the RSNA test set (Fig. \ref{fig3} (\textbf{b})), showing a more substantial discriminative capability of incorporating pseudo labeling (improved $\chi ^{2-\text{distance}}$ from 38.53 to 58.37). Finally, the ViT encoder obtained by MAE pre-training (\textbf{Stage 1}) surpasses DenseNet-121 (DN121), either pre-trained by MAE~\cite{xiao2023delving} or an advanced contrastive learning method (MoCo v2~\cite{chen2020improved}) on the RSNA dataset (Fig. \ref{fig3} (\textbf{c})).



%
%


%
%
%
%
\section{Conclusion}
We present $\AMAE$, an adaptation strategy of the pre-trained MAE for dual distribution anomaly detection in CXRs, which makes our method capable of more effectively apprehending anomalous features from unlabeled images. Experiments on the three CXR benchmarks demonstrate that $\AMAE$ is generalizable to different model architectures, achieving SOTA performance. As for the limitation, an adequate number of normal training images is still required, and we will extend our pseudo-labeling scheme in our future work for robust anomaly detection bypassing any training annotations. 

\bibliographystyle{splncs04}
\bibliography{paper-1205_Refs}
\end{document}